\documentclass{article}

    \PassOptionsToPackage{numbers, compress}{natbib}

     \usepackage[preprint]{neurips_2024}

\usepackage[utf8]{inputenc} 
\usepackage[T1]{fontenc}    
\usepackage{hyperref}       
\usepackage{url}            
\usepackage{booktabs}       
\usepackage{amsfonts}       
\usepackage{nicefrac}       
\usepackage{microtype}      
\usepackage{xcolor}         
\usepackage{booktabs} 
\usepackage{resizegather}
\usepackage{algorithm}
\usepackage{algorithmic}
\usepackage{amsmath}
\usepackage{amssymb}
\usepackage{mathtools}
\usepackage{amsthm}
\usepackage{graphicx}
\usepackage{caption}
\usepackage{subcaption}

\title{Single Domain Generalization with Model-aware Parametric Batch-wise Mixup}

\author{%
	Marzi Heidari\quad \textnormal{and}\quad Yuhong Guo \\
  School of Computer Science, Carleton University, Ottawa, Canada
}

\begin{document}

\maketitle

\begin{abstract}
Single Domain Generalization (SDG) remains a formidable challenge in the field of machine learning, 
particularly when models are deployed in environments that differ significantly from their training domains.  
In this paper, we propose a novel data augmentation approach, 
named as Model-aware Parametric Batch-wise Mixup (MPBM), 
to tackle the challenge of SDG. 
MPBM deploys adversarial queries generated with stochastic gradient Langevin dynamics,
and produces model-aware augmenting instances with a parametric batch-wise mixup generator network
that is carefully designed through an innovative attention mechanism. 
By exploiting inter-feature correlations, the parameterized mixup generator 
introduces additional versatility in combining features across a batch of instances, 
thereby enhancing the capacity to generate highly adaptive and informative synthetic instances 
for specific queries. 
The synthetic data produced by this adaptable generator network, 
guided by informative queries, is expected to significantly enrich 
the representation space covered by the original training dataset 
and subsequently enhance the prediction model's generalizability 
across diverse and previously unseen domains.
To prevent excessive deviation from the training data, 
we further incorporate a real-data alignment-based adversarial loss into 
the learning process of MPBM, regularizing any tendencies toward undesirable expansions.  
We conduct extensive experiments on several benchmark datasets. 
The empirical results demonstrate that by augmenting the training set
with informative synthesis data, our proposed MPBM method
achieves the state-of-the-art performance for single domain generalization. 
\end{abstract}

\section{Introduction}

Domain generalization (DG) is a critical challenge in machine learning, aiming to train a model on data from 
observed source domains and deploy it successfully on previously unseen target domains. 
This task is particularly demanding because it requires the model to be robust and adaptable to new, potentially divergent environments without access to the target domain during training \cite{blanchard2011generalizing, muandet2013domain}. Despite substantial progress, traditional DG approaches often struggle with significant domain shifts that are not adequately represented in the training data \cite{wang2021cross, gulrajani2020search},
leading to poor generalization to new, unseen environments. 
These challenges intensify in Single Domain Generalization (SDG) scenarios,  
where training data is solely sourced from a single domain,
thereby further constraining the diversity and richness of the available training data.
Therefore, there is a pressing need for innovative strategies that can enhance the robustness and adaptability of 
models trained on any source domain. 
Ensuring reliable performance across a broad spectrum of unseen scenarios requires the development of techniques that mitigate overfitting and promote generalization beyond the training domain \cite{volpi2018generalizing, qiao2020learning}.

One promising approach to address this challenge is the use of data augmentation techniques, which aim to increase the diversity of training data by generating synthetic examples. Among these techniques, Mixup has garnered considerable attention for its effectiveness in improving model generalization. 
Mixup generates new training samples by linearly interpolating between pairs of existing examples, thereby creating a more diverse set of training conditions \cite{zhang2018mixup}. 
Although approaches such as AutoMixUp \cite{liu2022automix} have attempted to parameterize the learning of the mixup process, they remain constrained to linear combinations of a pair of samples.  
Despite their efficacy, traditional linear pairwise combination based Mixup methods 
do not fully exploit the potential of generating versatile 
synthetic data to challenge and refine the model's learning process. 
Consequently, these methods often generate samples that are too similar to the original data, 
limiting their ability to prepare the model for the diverse and potentially adversarial conditions it might encounter post-training.

In this paper, we propose a novel approach 
termed Model-aware Parametric Batch-wise Mixup (MPBM) to synthesize 
informative and adaptive instances to enrich 
training data for single domain generalization. 
Unlike traditional Mixup approaches, which are constrained to parameterless linear interpolations 
between pairs of instances independent of the prediction model, 
the proposed MPBM significantly enhances the capacity and flexibility of mixup data synthesis in three key ways.
{\em Firstly}, MPBM facilitates mixup data synthesis across entire batches of instances rather than just pairs, 
broadening its synthesis and outcome spaces. 
{\em Secondly}, by devising a parameterized attention module for learning mixup coefficients
with extra heterogeneity along the feature dimension, 
MPBM can dynamically adapt its augmentation mechanism to generate very diverse
synthetic instances, 
enhancing the richness and adaptability of the synthesized data.
{\em Thirdly}, by guiding the attention-based mixing up process with 
adversarial queries generated from the prediction model 
using stochastic gradient Langevin dynamics, 
we enable model-aware instance synthesis, 
augmenting the training set 
with particularly informative samples, 
and bolstering the prediction model's generalization capabilities.
Additionally, we regulate the learning process of MPBM through an adversarial training regime, 
mitigating the risk of generating synthetic data 
that deviates excessively from the training dataset
and thereby averting undesirable negative expansions.
To evaluate the effectiveness of our proposed method in enriching the training data and 
enhancing the generalizability of the trained prediction model, 
we conduct experiments on multiple benchmark datasets for single-domain generalization. 
Our PMBM approach demonstrates remarkable improvements in generalization performance compared to baseline methods, 
highlighting its efficacy in data augmentation and support for single-domain generalization tasks.

\section{Related work}

\subsection{Mixup Methods}
Mixup methods have gained significant traction in recent years as a powerful data augmentation technique to enhance the generalization and robustness of machine learning models. The foundational work by Zhang et al. \cite{zhang2018mixup} introduced Mixup, which generates synthetic training examples through linear interpolations of pairs of examples and their labels: 
$\mathbf{x}_{\text{mix}}=\lambda \mathbf{x}_i + (1-\lambda)\mathbf{x}_j$ 
and $\mathbf{y}_{\text{mix}}=\lambda \mathbf{y}_i + (1-\lambda)\mathbf{y}_j$. 
This simple yet effective approach has been widely adopted and has inspired numerous variants and extensions.
Building upon the original Mixup concept, CutMix \cite{yun2019cutmix} advances the idea by combining patches from different images rather than entire images, thereby preserving more local information and improving model performance. 
AugMix \cite{hendrycks2019augmix} augments images using a combination of simple image transformations and mixes the resulting images to create robust augmentations. 
Manifold Mixup \cite{verma2019manifold} performs mixup operations in the feature space rather than the input space, leading to smoother decision boundaries and better generalization.
AutoMixup \cite{liu2022automix} parameterizes the mixup process, 
providing a more automated way to generate mixup samples. 
Mixup has also been used to address various specific challenges, 
such as semi-supervised learning (SSL) and domain generalization. 
MixMatch \cite{berthelot2019mixmatch} leverage mixup data 
to enhance SSL from limited labeled data. 
MixStyle \cite{zhou2020domain} mixes the feature statistics of training samples across multiple source domains 
to increase the domain diversity of the source domains. 
Despite these advancements, most existing methods rely on linear combinations of pairwise samples, 
limiting their ability of 
generating diverse instances
for challenging and refining the model's learning process.

Our proposed method, Model-aware Parametric Batch-wise Mixup (MPBM), 
advances the mixup technique in key aspects and 
addresses these limitations by 
parametrizing a query-guided feature-wise mixup process 
using learnable attention mechanisms. 
This approach adaptively generates more complex and informative synthetic examples
based on both the current prediction model and the training set, 
thereby enhancing model robustness and generalization capabilities.

\subsection{Single Domain Generalization}
The concept of Single Domain Generalization (SDG) centers on 
the task of learning generalizable models using only a single source domain, without any exposure to target domain distributions. 
Unlike traditional domain generalization which benefits from multiple source domains to enhance model robustness and generalizability, 
SDG must rely solely on a single source domain to overcome potential distribution shifts that might be encountered in unseen target domains,
thereby posing much more difficult challenges. 
Current methodologies in SDG fall into three main categories,
primarily centering on enriching the training data through data augmentation. 
Firstly, traditional data augmentation techniques are employed to bolster in-domain robustness.
Seminal works in this category include improved augmentation strategies \cite{devries2017improved}, AugMix \cite{hendrycks2019augmix}, and AutoAugment \cite{Cubuk_2020_CVPR_Workshops}, 
which primarily enhance sample variability within the source domain 
to improve out-of-domain generalization. 
Lian et al. \cite{lian2021geometry} extended this notion by introducing geometric transformations aimed at enriching the diversity of training samples. 
ACVC \cite{Cugu_2022_CVPR} samples transformations from a pool of visual corruptions to simulate distinct training domains and enforces visual attention consistency between the original and corrupted samples.
The second category introduces adversarial data augmentation approaches, which creatively manipulate either pixel space or latent features to mimic domain variability. Volpi et al. \cite{volpi2018generalizing} and Zhao et al. \cite{Long2020Maximum} pioneered augmenting images directly in the pixel space, aiming to conjure worst-case domain shifts. Subsequently, work by  Zhang et al. \cite{zhang2023adversarial} advances this concept by perturbing latent feature statistics, yet consistently generating extensive domain shifts remains a challenge. 
Adversarial AutoAugment \cite{zhang2019adversarial} formulates augmentation policies through adversarial learning 
to enhance in-domain generalization.
The third category leverages generative models to synthesize training data. Innovations by Qiao et al. \cite{qiao2020learning}, Wang et al. \cite{wang2021learning}, and Li et al. \cite{li2021progressive} utilize generative adversarial networks (GANs) and variational autoencoders (VAEs) to generate diverse, albeit domain-constrained, samples. MCL \cite{chen2023meta} first simulates domain shift with an auxiliary target domain, then analyzes its causes, and finally reduces the shift for model adaptation. These methods produce synthetic data that, while stylistically varied, still reflect inherent characteristics of the source domain.

\begin{figure*}[t]
  \centering
 \includegraphics[width= 1\textwidth]{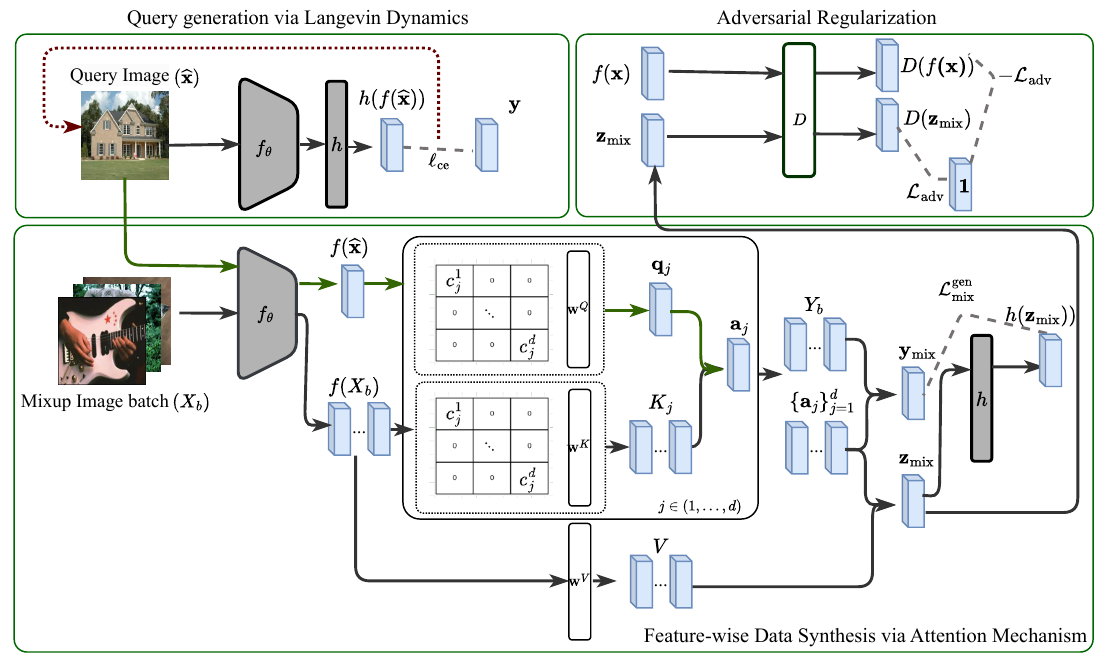} 
	 \caption{Overview of the training process for the mixup generator. The process begins with a pre-trained feature extractor $f_{\theta}$ and classifier $h_{\psi}$, whose parameters ($\theta$ and $\psi$) remain frozen during the training of the Mixup Generator Network (MGN) $g_{\phi}$. Query image inputs undergo Langevin Stochastic Query Augmentation. The Entry-Wise Feature Attention Mechanism captures intricate interactions among feature dimensions by incorporating a correlation matrix, facilitating precise feature mixups. The mixup generation loss $\mathcal{L}_{\text{mix}}^{\text{gen}}$ is being optimized for the generated mixup samples. The Adversarial Mixup Generation component aligns generated mixup features with real data through an adversarial training framework, optimizing the adversarial loss $\mathcal{L}_{\text{adv}}(\phi, \omega)$.
	}
   \label{fig:method}
   \vskip -0.1 in
\end{figure*}

\section{Method}
Given a labeled dataset $\mathcal{D} = \{(\mathbf{x}_i, \mathbf{y}_i)\}_{i=1}^N$ from a source domain
such that $\mathbf{x}_i\in\mathcal{X}$ and $\mathbf{y}_i\in\mathcal{Y}$,
the primary objective of Single Domain Generalization (SDG) is to learn a prediction model $(h \circ f): \mathcal{X} \rightarrow \mathcal{Y}$, 
comprising a feature extractor $f: \mathcal{X}\rightarrow\mathcal{Z}$ 
parameterized by $\theta$ and a classifier $h: \mathcal{Z}\rightarrow\mathcal{Y}$ 
parameterized by $\psi$,
that can generalize well on 
an unseen target domain denoted as $\mathcal{D}^t = \{(\mathbf{x}_i^t, \mathbf{y}_i^t)\}_{i=1}^{N^t}$,
which is inaccessible during the training phase. 

In this section, we present the proposed 
Model-aware Parametric Batch-wise Mixup (MPBM) approach 
for generating effective augmenting instances to enhance the generalizability 
of prediction models in the SDG scenario. 
The key of MPBM lies in the development of an attention mechanism capable of learning a feature-wise instance mixup generator from a batch of training samples. 
This innovation expands the data augmentation space, fostering diversity and breadth. 
Moreover, it guides the mixup process by utilizing informative queries generated from the prediction model through stochastic gradient Langevin dynamics.
To avoid excessive deviation and mitigate negative data augmentation, 
we further integrate an adversarial loss to regulate the learning of the mixup generator. 
The resulting synthetic data can then be readily employed to augment supervised learning 
on the original training set, thereby enhancing the model's capacity to generalize to unseen domains.
The details are elaborated below.

\subsection{Model-aware Parametric Batch-wise Mixup}
Towards the goal of improving the generalizability of the prediction model,
we propose to conduct model-aware data augmentation
by generating synthetic data based on the current prediction model. 

Initially, given the labeled training set $\mathcal{D}$ in the source domain, 
we apply the standard supervised learning to learn the prediction model 
by minimizing the following cross-entropy loss: 
\begin{equation}
\label{eq:loss-sup}
\mathcal{L}_{\text{sup}}(\theta,\psi)=\mathbb{E}_{(\mathbf{x}, \mathbf{y}) \in \mathcal{D}}[\ell_{ce}(h_\psi(f_\theta(\mathbf{x}), \mathbf{y})]
\end{equation}
where $\ell_{ce}$ denotes the cross-entropy loss, effectively quantifying the discrepancy between the predicted and actual labels, and facilitating the learning of the feature extractor $f$ and classifier $h$. 

\subsubsection{Feature-wise Data Synthesis via Attention Mechanism}
Unlike traditional mixup approaches that conduct parameterless linear
interpolations between pairs of instances, 
we devise an innovative Mixup Generator $g_\phi$ parametrized with $\phi$, 
to synthesize diverse instances in the feature space $\mathcal{Z}$
over a batch of $N_b$ training samples ($N_b\geq 2$) based on an attention mechanism~\cite{Attention2017},
while allowing extra mixing heterogeneity along the feature dimension.

Specifically, given a query instance $\widehat{\mathbf{x}}_q$, 
the mixup generator $g_\phi$ deploys 
an attention mechanism to 
mix up a new instance $(\mathbf{z}_{\text{mix}},\mathbf{y}_{\text{mix}})$
in the feature space such that $\mathbf{z}_{\text{mix}}\in\mathcal{Z}$
by integrating information
from a training batch $(X_b, Y_b)$ that includes $N_b$ training instances
based on their affinities with the query instance---{\em i.e.}, attention scores.
To compute insightful affinity information that are closely relevant to the prediction model,
we deploy the attention score computation in the extracted feature space $\mathcal{Z}\subseteq\mathbb{R}^d$. 
Although standard attention mechanisms produce a single attention score
for all the features of each instance in the batch, 
we further offer an extra degree of versatility 
by allowing different attention scores for different features in each instance. 
Without increasing the number of parameters in the generator $g_\phi$ and hence the computational complexity, 
we achieve this flexibility by leveraging the interrelationships among the features 
encoded by a feature correlation matrix 
$C \in \mathbb{R}^{d \times d}$ computed from the training set $\mathcal{D}$,
where $d$ denotes the size of the feature dimension. 
The $j$-th row of $C$, denoted as $\mathbf{c}_j$, is a 
probability distribution vector delineating correlations between the $j$-th feature and the other features. 
It is computed by normalizing the vector of Pearson correlation \cite{pearson1895vii} coefficients 
computed between pairs of features. 
To facilitate subsequent presentation, we denote the diagonal matrix with $\mathbf{c}_j$
as the diagonal vector as $C_j\in\mathbb{R}^{d\times d}$, such that $C_j=\text{diag}(\mathbf{c}_j)$. 

Concretely, to compute the attention scores for the $j$-th feature over the batch of $N_b$ instances, 
we compute the query vector $\mathbf{q}_j\in \mathbb{R}^{1 \times d}$, 
the key matrix $K_j\in\mathbb{R}^{N_b\times d}$
and the value matrix $V\in\mathbb{R}^{N_b\times d}$ 
under the parametric attention mechanism as follows: 
\begin{equation}
\label{eq:query-key}
	\mathbf{q}_j = {\mathbf{z}_q} C_j W^Q, 
	\quad K_j = Z_b C_j W^K, 
	\quad V = Z_b W^V, 
\end{equation}
where $\mathbf{z}_q=f_{\bar{\theta}}({\widehat{\mathbf{x}}_q})\in\mathbb{R}^{1\times d}$ 
and $Z_b=f_{\bar{\theta}}(X_b)\in\mathbb{R}^{N_b\times d}$
are the extracted feature vector and feature matrix of the query instance
and the batch of training instances, respectively; 
${\bar{\theta}}$ denotes the stop-gradient version of $\theta$; 
the weight matrices $W^Q$, $W^K$, and ${W}^V$ in the size of $d\times d$
form the parameters $\phi$ of the mixup generator $g_\phi$. 
By introducing the feature correlation matrix $C$ into the query and key computation, 
this allows us to compute different attention scores for each individual feature 
with attention mechanisms operated in the whole feature space.
The vector of attention scores $\mathbf{a}_j \in \mathbb{R}^{1 \times N_b}$ 
over the $N_b$ instances 
for the $j$-th feature is then computed through a softmax operation over 
the scaled dot-products between the query and keys,
which capture the affinities between them: 
\begin{equation}
\label{eq:attention}
	\mathbf{a}_j = \text{softmax}\left(\big(\mathbf{q}_j K_j^\top\big)/\sqrt{d}\right).
\end{equation}
In parallel, we can compute all the $d$ vectors of attention scores for 
the $d$ features. 
These scores are crucial for selectively aggregating feature values across the $N_b$ instances, 
forming a new synthetic instance 
$\mathbf{z}_{\text{mix}}$ 
with enriched feature representations 
via multiplication with the value matrix:
\begin{equation}
\label{eq:z-aug}
\mathbf{z}_{\text{mix}} = \left[\mathbf{a}_1 \mathbf{v}_1, \cdots,\mathbf{a}_j \mathbf{v}_j,\cdots,\mathbf{a}_d \mathbf{v}_d\right],
\end{equation}
where $\mathbf{v}_j$ denotes the $j$-th column
of the value matrix $V$. 
Simultaneously, the corresponding synthetic label vector 
$\mathbf{y}_{\text{mix}}$ can be generated by applying a softmax function to the average of the attention scores across feature dimension $d$, and subsequently aggregating the label matrix $Y_b$:
\begin{equation}
\label{eq:y-aug}
\mathbf{y}_{\text{mix}} = \text{softmax}\left(\frac{1}{d} \sum\nolimits_{j=1}^d \mathbf{a}_j\right) Y_b.
\end{equation}
This mixup data generation process can be summarized as
a parametric mixup generator $g_\phi$:
\begin{equation}
\label{eq:mixup-generate}
	(\mathbf{z}_{\text{mix}},\mathbf{y}_{\text{mix}}) 
	= g_\phi(\widehat{\mathbf{x}}_q, X_b, Y_b).
\end{equation}
For convenience, we use 
$g_\phi^z$ to denote a partial $g_\phi$ that 
only outputs the synthesized vector $\mathbf{z}_{\text{mix}}$.

\subsubsection{Adversarial Query Generation via Stochastic Gradient Langevin Dynamics}

The mixup generator introduced above synthesizes instances by aggregating a training batch 
based on their parametric affinities with the query instance. 
In this process, 
the query instance assumes a pivotal role in guiding the mixup process and influencing its outcomes. 
Therefore, to purposefully enhance the generalization capacity of the prediction model, 
we propose to generate adversarial query instances in a model-aware manner. 
The idea is to produce queries capable of diverging from existing training instances 
under the current prediction model, 
thereby expanding the feature representation space in a way that 
can bolster the model's generalizability.

This adversarial query generation process
is realized using the stochastic gradient Langevin Dynamics
\cite{Langevin2011,braun2024deep}. 
Specifically, given a labeled instance 
$(\mathbf{x}_i, \mathbf{y}_i)$ randomly selected from the training set $\mathcal{D}$
as a query seed,
we generate a query instance $\widehat{\mathbf{x}}_q$ 
from it by adversarial maximizing the cross-entropy loss
of the prediction model in a stochastic manner with the following Markov chain of updates: 
\begin{equation}
	\label{eq:langevin-step}
	\widehat{\mathbf{x}}_i^{t+1} = \widehat{\mathbf{x}}_i^{t} 
	+ \eta(t) \nabla_{\widehat{\mathbf{x}}_i} \ell_{ce}(h_\psi(f_\theta(\widehat{\mathbf{x}}_i^t)), \mathbf{y}_i) + \sqrt{2 \eta(t)} \mathbf{\epsilon}_i^{t},
\end{equation}
where $\mathbf{\epsilon}_i^{t} \sim \mathcal{N}(0, \mathbf{I})$
for $t = 1, \cdots, T$, and  
$\eta(t)$ is the stepsize at time-step $t$;
$\nabla_{\widehat{\mathbf{x}}_i} \ell_{ce}(h_\psi(f_\theta(\widehat{\mathbf{x}}_i^t)), \mathbf{y}_i)$ 
denotes the gradient of the cross-entropy loss of the prediction model 
on the labeled instance pair $(\widehat{\mathbf{x}}_i^t, \mathbf{y}_i)$ 
with respect to the input data. 
The stochastic Gaussian noises are injected to prevent collapse and increase variability. 
The update process starts with $\widehat{\mathbf{x}}_i^{1} = \mathbf{x}_i$ and generates
an adversarial query instance $\widehat{\mathbf{x}}_q=\widehat{\mathbf{x}}_i^{T+1}$ in $T$ steps. 
The standard stochastic gradient Langevin dynamics \cite{Langevin2011} 
is designed to produce samples from a probability density using only the gradients
of the log probability density with a Markov chain of updates.
Here we adapt it to produce query samples from a distribution that maximizes 
the cross-entropy loss under the current prediction model.  
We denote this model-aware adversarial query generation process as the following function:
$\widehat{\mathbf{x}}_q = \widehat{g}_q(\mathbf{x}_i)$.

\subsubsection{Learning Mixup Generator with Adversarial Regularization}
To prevent arbitrary diversity that is irrelevant for prediction,
the training of the mixup generator $g_\phi$ is primarily driven by 
the fitness of the produced synthetic instances under the prediction model,
as encoded by the following loss function:
\begin{equation}
\label{eq:loss-aug}
\mathcal{L}_{\text{mix}}^{\text{gen}} (\phi)= \mathbb{E}_{(X_b, Y_b) \sim \mathcal{D},{\mathbf{x}}_i \sim \mathcal{D} }
	[\ell_{ce}\big(h_{\bar{\psi}}(g_{\phi}(\widehat{g}_q(\mathbf{x}_i),X_b, Y_b))\big)],
\end{equation}
where $\bar{\psi}$ denote the stop-gradient version of $\psi$.

Moreover, to mitigate the risk of excessive deviation and negative data expansion, 
we further regulate the learning process of the mixup generator through an adversarial training regime.
This is implemented through the following real-data alignment based adversarial loss: 
\begin{equation}
\label{eq:adv-loss}
\mathcal{L}_{\text{adv}}(\phi, \omega) = 
\mathbb{E}_{\mathbf{x} \sim \mathcal{D}}[\log D_\omega(f_{\bar{\theta}}(\mathbf{x}))] 
	+ \mathbb{E}_{X_b \sim \mathcal{D},\mathbf{x}_i \sim \mathcal{D} }
	[\log(1 - D_\omega(g_\phi^z(\widehat{g}_q(\mathbf{x}_i),X_b, \cdot)))],
\end{equation}
where $D_\omega$ is a discriminator parameterized by $\omega$, designed to differentiate between feature vectors of real data and those generated through the mixup process. 
The learning is conducted as a minimax adversarial game such as: 
$\min_\phi \max_\omega \mathcal{L}_{\text{adv}}(\phi, \omega)$,
where the discriminator tries to maximize the objective in Eq.(\ref{eq:adv-loss}),
distinguishing the feature distribution of the synthetic data from that of the real data,
while the mixup generator $g_\phi$ is learned to minimize the objective
to align the two distributions.

By integrating the prediction loss over the synthetic instances in Eq.(\ref{eq:loss-aug})
and the adversarial alignment loss in Eq.(\ref{eq:adv-loss}) together, 
we learn the parametric mixup generator $g_\phi$ 
through the following min-max optimization: 
\begin{equation}
\min_{\phi} \max_{\omega}\mathcal{L}_\text{gen}= \mathcal{L}_{\text{mix}}^{\text{gen}} + \lambda_{\text{adv}} \mathcal{L}_{\text{adv}},
\end{equation}
where $\lambda_{\text{adv}}$ is a hyperparameter that controls the contribution of the adversarial loss.
A very small value of $\lambda_{\text{adv}}$ will diminish the effectiveness of the regularization 
in mitigating negative data generation, 
whereas a large value of $\lambda_{\text{adv}}$ 
may overly constrain the mixup generator, limiting its ability to effectively expand the feature distribution coverage.
The training process of the mixup generator is also illustrated in Figure \ref{fig:method}.

\begin{algorithm}[t]
  \caption{Training Algorithm for MPBM}
  \begin{algorithmic} 
 \STATE \textbf{Input}: $\mathcal{D}$; Mixup Generator $g_{\phi_0}$, pre-trained $f_{\theta_0}$ and  $h_{\psi_0}$, Discriminator $D_{\omega_0}$
\STATE \textbf{Output}: Trained  prediction model $f_\theta,h_\psi$, Mixup Generator $g_{\phi}$  , Discriminator model $D_{\omega}$\\[1ex]
\STATE \textbf{Set} $\mathcal{D}_{\text{mix}}=\{\}$
\FOR{{iteration i = 1} {\bf to} I} 
    \STATE Fix parameters of $f_\theta$  and $h_\psi$ 
    \STATE Calculate correlation matrix $C$ using the features of training data $\{ f_{\bar{\theta}}(\mathbf{x}_i)\}_{i=1}^N$
\STATE Sample the query set $\{\mathbf{x}_i\}_{i=1}^m$ from $\mathcal{D}$  
	  and generate $\mathcal{Q}=\{\widehat{\mathbf{x}}_q^i = \widehat{g}_q(\mathbf{x}_i)\}_{i=1}^m$ through Eq.(\ref{eq:langevin-step})
	  \STATE Sample mixup base batches $\mathcal{B}=\{(X_b,Y_b)_i\}_{i=1}^m$ from $\mathcal{D}$ 

\FOR{{iteration j = 1} {\bf to} J} 
	  \STATE Generate mixup data $\mathcal{M}=\{(z_{\text{mix}}^i, y_{\text{mix}}^i) \}_{i=1}^m$ 
	  from $\mathcal{Q}$ and $\mathcal{B}$ with $g_\phi$ in Eq.(\ref{eq:mixup-generate})  
	  \STATE Sample a batch of real data $X_{B}$ from $\mathcal{D}$ and 
	  calculate $\mathcal{L}_{\text{adv}}$ using Eq.(\ref{eq:adv-loss}) on $X_{B}$ and $\mathcal{M}$
   \STATE Update $\omega$ using gradient ascend on $\mathcal{L}_{\text{adv}}$ 
	  \STATE Calculate cross-entropy loss $\mathcal{L}_{\text{mix}}^\text{gen}$ using Eq.(\ref{eq:loss-aug}) on $\mathcal{M}$ 
 
    \STATE Update parameters of $g_\phi$ using gradient descent on $ \mathcal{L}_{\text{gen}} =\mathcal{L}_{\text{mix}}^{\text{gen}} + \lambda_{\text{adv}}  \mathcal{L}_{\text{adv}}$ 
    \ENDFOR
    \STATE Fix parameters of $g_\phi$  and $D_\omega$
   \STATE Generate mixup data $\{(z_{\text{mix}}^i, y_{\text{mix}}^i) \}_{i=1}^m$ 
	  from $\mathcal{Q}$ and $\mathcal{B}$ with current $g_\phi$ and add them to $\mathcal{D}_\text{mix}$
    \FOR{{iteration t = 1} {\bf to} T} 
	  \STATE Calculate $\mathcal{L}_\text{sup}$ using Eq.(\ref{eq:loss-sup}) 
	       on a batch $\mathcal{D}_b$ randomly sampled from $\mathcal{D}$
	  \STATE Calculate $\mathcal{L}_{\text{mix}}^{\text{tr}}$ on  $\mathcal{D}_{\text{mix}}$ using Eq.(\ref{eq:aug-mixloss})
    \STATE Update  $\theta$ and $\psi$ using gradient descent on $\mathcal{L}_{\text{tr}} = \mathcal{L}_\text{sup} + \lambda_\text{mix} \mathcal{L}_{\text{mix}}^{\text{tr}}$
    
    \ENDFOR
    
\ENDFOR
  \end{algorithmic}
  \label{alg:train}
\end{algorithm}

\subsection{Data Augmentation for Model Generalization}
By randomly selecting query seeds and batches of training samples,
we can deploy the learned mixup generator $g_\phi$ to generate a set of 
synthetic instances $\mathcal{D}_{\text{mix}}$,
which can be incorporated into the training regime to improve the prediction model
through the following augmentation loss:
\begin{equation}
\label{eq:aug-mixloss}
\mathcal{L}_{\text{mix}}^{\text{tr}}(\theta,\psi)= \mathbb{E}_{(\mathbf{z}_{\text{mix}}, \mathbf{y}_{\text{mix}}) \in \mathcal{D}_{\text{mix}}}[\ell_{ce}(h_\psi(\mathbf{z}_{\text{mix}}), \mathbf{y}_\text{mix})].
\end{equation}

The data augmentation is progressively conducted, starting with 
an empty augmentation dataset $\mathcal{D}_{\text{mix}}$ and being iteratively populated 
through a structured process. 
Once the augmentation data is populated, we fine-tune the prediction model 
by integrating both the real training set $\mathcal{D}$ and  
the augmentation dataset $\mathcal{D}_{\text{mix}}$ through the following combined loss minimization:
\begin{equation}
\min_{\theta, \psi} \mathcal{L}_{\text{tr}} = \mathcal{L}_\text{sup} + \lambda_\text{mix} \mathcal{L}_{\text{mix}}^{\text{tr}},
\end{equation}
where $\lambda_\text{mix}$ is a trade-off hyperparameter. 
By gradually expanding the augmentation dataset along with the progressive improvement of the prediction model, 
we expect to increase the model's exposure to a wider array of data variations in a stable manner, 
fostering a more adaptable learning process and 
enhancing the model's generalizability to new and unseen data. 
The detailed training algorithm is outlined in Algorithm \ref{alg:train}.

\begin{table*}[t]
\centering
{\small
\caption{Classification accuracy and standard deviation(\%) comparison on the PACS dataset. Best results are in bold font.}
\label{tab:pacs-experiment}
\setlength{\tabcolsep}{4pt}
\begin{tabular}{c|cccccccc|c}
\hline
Target & MixUp & CutOut & ADA & ME-ADA & AugMix & RandAug & ACVC& L2D  & MPBM (Ours) \\ \hline
Art & 52.8 & 59.8 & 58.0 & 60.7 & 63.9 & 67.8 & 67.8 & 67.6 & $\mathbf{68.5}_{(1.3)}$  \\
Cartoon & 17.0 & 21.6 & 25.3 & 28.5 & 27.7 & 28.9 & 30.3 & 42.6 & $\mathbf{45.5}_{(1.2)}$  \\
Sketch & 23.2 & 28.8 & 30.1 & 29.6 & 30.9 & 37.0 & 46.4 & 47.1 & $\mathbf{49.7}_{(1.5)}$ \\
Avg. & 31.0 & 36.7 & 37.8 & 39.6 & 40.8 & 44.6 & 48.2 & 52.5 & $\mathbf{54.5}$ \\ \hline
\end{tabular}
}
\end{table*}
\begin{table*}[t]
\centering
{\small
\caption{Classification accuracy and standard deviation(\%) results on the four target domains SVHN, MNIST-M, SYN, and USPS, with MNIST as the source domain. Best results are in bold font.}
\setlength{\tabcolsep}{10pt}
\label{tab:digits}
\begin{tabular}{l|c|c|c|c|c}
\hline
Method & SVHN & MNIST-M & SYN & USPS & Avg. \\ \hline
 ERM \cite{koltchinskii2011oracle} & 27.8 & 52.7 & 39.7 & 76.9 & 49.3 \\
 CCSA \cite{motiian2017unified}& 25.9 & 49.3 & 37.3 & 83.7 & 49.1\\
JiGen \cite{carlucci2019domain} & 33.8 & 57.8 & 43.8 & 77.2 & 53.1\\
ADA \cite{volpi2018generalizing}& 35.5 & 60.4 & 45.3 & 77.3 & 54.6\\
 ME-ADA \cite{Long2020Maximum} & 42.6 & 63.3 & 50.4 & 81.0 & 59.3\\ 
M-ADA \cite{qiao2020learning} & 42.6 & 67.9 & 49.0& 78.5 & 59.5\\
AutoAug \cite{cubuk2018autoaugment} & 45.2 & 60.5 & 64.5 & 80.6 & 62.7 \\
RandAug \cite{cubuk2020randaugment} & 54.8 & 74.0 & 59.6 & 77.3 & 66.4 \\

 L2D \cite{wang2021learning}& 62.9 &  \textbf{87.3} & 63.7 & 84.0 & 74.5\\ 
  PDEN \cite{li2021progressive} & 62.2 & 82.2 & 69.4 & 85.3 & 74.8\\
    MCL\cite{chen2023meta} & 69.9 & 78.3 & 78.4 & 88.5 & 78.8 \\
    RSDA \cite{volpi2019addressing} & ${47.7}_{(4.8)}$ & ${81.5}_{(1.6)}$ & ${62.0}_{(1.2)}$ & ${83.1}_{(1.2)}$ & 68.5\\\hline 
       MPBM (Ours) & $\mathbf{70.9}_{(0.9)}$ & ${85.0}_{(0.5)}$ & $\mathbf{79.9}_{(0.8)}$ & $\mathbf{89.4}_{(0.4)}$ & $\mathbf{81.3}$ \\\hline
\end{tabular}}
\end{table*}
\begin{table*}[t]
\centering
{\small
\caption[DomainNet result]{Classification accuracy and standard deviation(\%) comparison on the DomainNet dataset. Best results are in bold font.}
\label{tab:domainnet}
\setlength{\tabcolsep}{4pt}
\begin{tabular}{c|cccccccc|c}
\hline
Target    & MixUp & CutOut & CutMix & ADA  & ME-ADA & RandAug & AugMix & ACVC          & MPBM (Ours)                                        \\ \hline
Painting  & 38.6  & 38.3   & 38.3   & 38.2 & 39.0   & 41.3    & 40.8   & 41.3          & $\mathbf{42.3}_{(0.3)}$ \\
Infograph & 13.9  & 13.7   & 13.5   & 13.8 & 14.0   & 13.6    & 13.9   & 12.9          & $\mathbf{14.8}_{(0.3)}$  \\
Clipart   & 38.0  & 38.4   & 38.7   & 40.2 & 41.0   & 41.1    & 41.7   & ${42.8}$ & $\mathbf{43.5}_{(0.4)} $ \\
Sketch    & 26.0  & 26.2   & 26.9   & 24.8 & 25.3   & 30.4    & 29.8   & 30.9          & $\mathbf{31.8}_{(0.3)}$ \\
Quickdraw & 3.7   & 3.7    & 3.6    & 4.3  & 4.3    & 5.3     & 6.3    & 6.6  & $\mathbf{7.1}_{(0.2)}$    \\
Avg.      & 24.0  & 24.1   & 24.2   & 24.3 & 24.7   & 26.3    & 26.5   & 26.9          & $\mathbf{27.9}$ \\ \hline
\end{tabular}}
\end{table*}
\begin{table*}[t]
\centering
{\small
\caption{Ablation Study Classification accuracy and standard deviation(\%) results on the four target domains SVHN, MNIST-M, SYN, and USPS, with MNIST as the source domain. Best results are in bold font.}
\label{tab:ablation}
\setlength{\tabcolsep}{12pt}
\begin{tabular}{l|c|c|c|c|c}
\hline
Method & SVHN & MNIST-M & SYN & USPS & Avg. \\ \hline
        MPBM (Ours) & $\mathbf{70.9}_{(0.9)}$ & $\mathbf{{85.0}}_{(0.5)}$ & $\mathbf{79.9}_{(0.8)}$ & $\mathbf{89.4}_{(0.4)}$ & $\mathbf{81.3}$ \\\hline
        $- \text{w/o } \mathcal{L}_\text{mix}^{\text{tr}}$ &${30.1}_{(2.4)}$ & ${53.1}_{(1.6)}$  & ${42.3}_{(2.1)}$  & ${81.3}_{(0.5)}$  & ${51.7}$ \\
                   $- \text{w/o } \mathcal{L}_\text{adv}$ &${68.1}_{(0.7)}$ & ${81.2}_{(1.0)}$  & ${75.7}_{(0.7)}$  & ${85.9}_{(0.8)}$  & $77.7$ \\
                 
                     $- \text{w/o } \mathcal{L}_\text{mix}^{\text{gen}}$ &${60.5}_{(1.1)}$ & ${75.2}_{(1.2)}$  & ${63.1}_{(1.9)}$  & ${80.1}_{(0.6)}$  & ${69.7}$ \\
                     $- \text{w/o  SGLD}$  &${69.1}_{(0.9)}$ & $82.6_{(0.8)}$  & $77.9_{(0.6)}$  & $87.4_{(0.6)}$  & $79.2$ \\ \hline

\end{tabular}}
\end{table*}

\section{Experiments}
\subsection{Experimental Setup}
\paragraph{Datasets} Our experimental evaluation utilizes three benchmark datasets. Digits benchmark serves as the foundation for digit classification, comprising five distinct datasets: MNIST \cite{lecun1998gradient}, MNIST-M \cite{ganin2015unsupervised}, SVHN \cite{netzer2011reading}, SYN \cite{ganin2015unsupervised}, and USPS \cite{denker1989neural}. Each dataset features an identical set of classes, specifically the digits 0 through 9. We designate MNIST as the source domain, with the remaining datasets serving as target domains for evaluation. 
PACS \cite{li2017deeper} encompasses four stylistically diverse domains: Art, Cartoon, Photo, and Sketch, all sharing seven common object categories. We select the Photo domain as the source while using the others as target domains. 
DomainNet \cite{peng2019moment}, the most demanding dataset in our study, spans six domains—Real, Infograph, Clipart, Painting, Quickdraw, and Sketch—and includes 345 object classes. The Real domain is used as the source, with the extensive class diversity and domain variability of the remaining domains presenting a significant challenge in our tests.
\paragraph{Implementation Details}
In our experiments, following the previous works \cite{wang2021learning}, we utilized LeNet as the architectural backbone for the Digits dataset, training on the first 10,000 MNIST images. All images were resized to $32 \times 32$ and converted to RGB format. We configured the experiment with 50 epochs, a batch size of 32, and an initial learning rate of $1e^{-4}$, which was reduced by a factor of 0.1 after 25 epochs.
For the PACS dataset, a pre-trained ResNet-18 on ImageNet was fine-tuned on the source domain with images resized to $224 \times 224$. The setup included 50 epochs, a batch size of 32, and a learning rate of 0.001 adjusted according to a cosine annealing scheduler.
The same ResNet-18 model was utilized for the DomainNet dataset experiments. However, given the complexity and diversity of this dataset, the experiments were extended to 200 epochs to allow deeper learning and better model convergence. We increased the batch size to 128 to process a larger volume of data per training iteration. The learning rate adjustment followed a similar cosine annealing pattern, tailored to sustain and enhance the model's performance over the extended training period.

All experiments were conducted with each setup replicated five times using different random seeds to ensure statistical reliability, and results were reported as the average accuracy with standard deviations. 
In configuring our method, we set the adversarial loss weight, $\lambda_{\text{adv}}$, and the mixing coefficient, $\lambda_{\text{mix}}$, each to 0.5, to balance their contributions to the model's learning process. We also specified $T$,  the number of Stochastic Gradient Langevin 
Dynamics steps, as 5. Additionally, The number
of instances in the mixup, $N_b$, was set to 5.

\subsection{Comparison Results}
We compare MPBM with several methods including MixUp \cite{zhang2018mixup}, CutOut \cite{devries2017improved}, 
CutMix \cite{yun2019cutmix}, 
AugMix \cite{hendrycks2019augmix}, ACVC \cite{Cugu_2022_CVPR}, ERM \cite{koltchinskii2011oracle}, CCSA \cite{motiian2017unified}, JiGen \cite{carlucci2019domain}, ADA \cite{volpi2018generalizing}, ME-ADA \cite{Long2020Maximum}, M-ADA \cite{qiao2020learning}, AutoAug \cite{cubuk2018autoaugment}, RandAug \cite{cubuk2020randaugment}, RSDA \cite{volpi2019addressing}, L2D \cite{wang2021learning}, PDEN \cite{li2021progressive}, and MCL \cite{chen2023meta}. using LeNet, ResNet-18 as the backbone network.

Table \ref{tab:pacs-experiment} reports the comparison results on the PACS dataset with a ResNet-18 backbone network.
Our proposed method, MPBM, demonstrates substantial improvements in classification accuracy across various target domains in the PACS dataset. MPBM achieves the highest average accuracy of 54.5\%, outperforming the second-best method L2D by a substantial margin of 2\%. Specifically, it excels in the "Art" domain with 68.5\%, surpassing ACVC; in the "Cartoon" domain with 45.5\%, significantly higher than L2D with a margin of 2.9\%; and in the "Sketch" domain, outpacing L2D the second best method by a significant margin of 2.6\%.

Table \ref{tab:digits} presents the comparative results on the digits dataset using LeNet as the backbone network. Our proposed method demonstrates superior performance, achieving the highest average classification accuracy of 81.3\%, surpassing the method with the second-best highest average, MCL, across all four target domains. Notably, our method attains a 1.0\% improvement over MCL on SVHN, a significant 6.7\% improvement on MNIST-M, a 1.5\% higher accuracy on SYN, and a 0.9\% gain in performance on USPS compared to MCL.

Table \ref{tab:domainnet} presents the comparative performance results on the DomainNet dataset, utilizing ResNet-18 as the backbone network. Our proposed method demonstrates superior performance, surpassing the second-best approach, ACVC, across all five target domains. Notably, our method achieves the highest average classification accuracy of 27.9\%, marking a significant improvement. Specifically, we observe a notable 1.9\% improvement margin in the Infograph domain, underscoring the efficacy of our approach in enhancing classification accuracy in challenging scenarios. These findings underscore the vital role of each component analyzed in our ablation study, confirming their significant contributions toward achieving robust and generalizing feature representations.

These remarkable gains across diverse domains illustrate the robustness of MPBM, reinforcing its capacity to adapt to varied environments and substantiating its effectiveness in generating synthetic samples that significantly enhance model generalization.

\subsection{Ablation Study}
We conducted an ablation study to investigate the contribution of different components of MPBM
on digits dataset by using LeNet as the backbone network. 
In particular, we compared the full model MPBM with the following variants:
(1) ``$\; - \text{w/o } \mathcal{L}_\text{mix}^{\text{tr}}$ ", which excludes mixup data from training;
(2) ``\; -$\text{w/o } \mathcal{L}_\text{adv}$", which adversarial regularization from training Mixup Generator; 
(3) ``$- \text{w/o } \mathcal{L}_\text{mix}^{\text{gen}}$", which drops the generation classification loss from training  Mixup Generator;
and
(4) ``$\; - \text{w/o  SGLD}$", which drops the Stochastic Gradient Langevin Dynamics from query samples;
The comparison results are reported in Table \ref{tab:ablation}.

From the results, it is evident that incorporating generated mixup data significantly enhances the model's performance. Specifically, the exclusion of the mixup training loss $\mathcal{L}_\text{mix}^{\text{tr}}$ results in the most substantial performance drop, highlighting the importance of mixup data in the training process. The absence of the adversarial loss $\mathcal{L}_\text{adv}$ also leads to a noticeable decline in accuracy, underscoring the role of adversarial training in refining the generated features. Additionally, the removal of the generation classification loss $\mathcal{L}_\text{mix}^{\text{gen}}$ and Stochastic Gradient Langevin Dynamics both contribute to performance degradation, demonstrating their significance in the overall framework. 
These findings confirm the critical contributions of each component in achieving robust and generalizing feature representations.
\begin{figure}[t]
\centering
\begin{subfigure}{0.23\textwidth}
\centering

\includegraphics[width = \textwidth, height=1.3in]{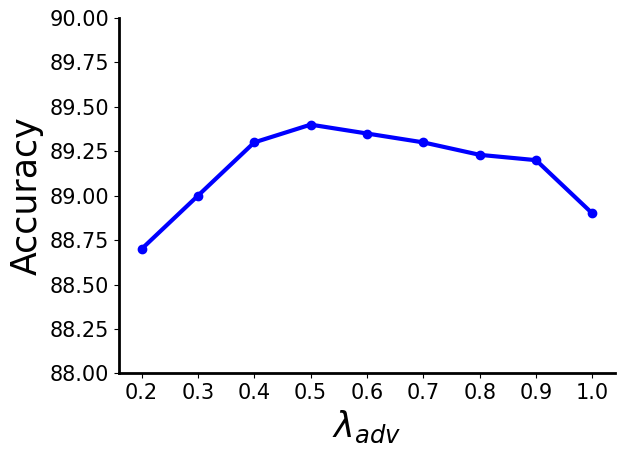}
\caption{$\lambda_{\text{adv}}$}
\end{subfigure}
\begin{subfigure}{0.23\textwidth}
\centering
\includegraphics[width = \textwidth, height=1.2in]{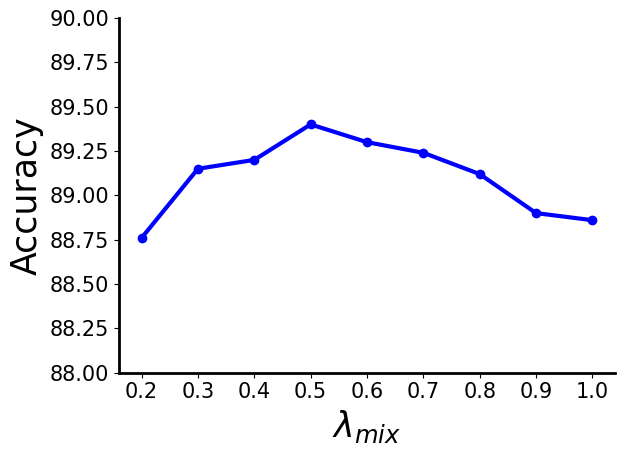}

\caption{$\lambda_{\text{mix}}$}
\end{subfigure}
\begin{subfigure}{0.23\textwidth}
\centering
\includegraphics[width = \textwidth, height=1.2in]{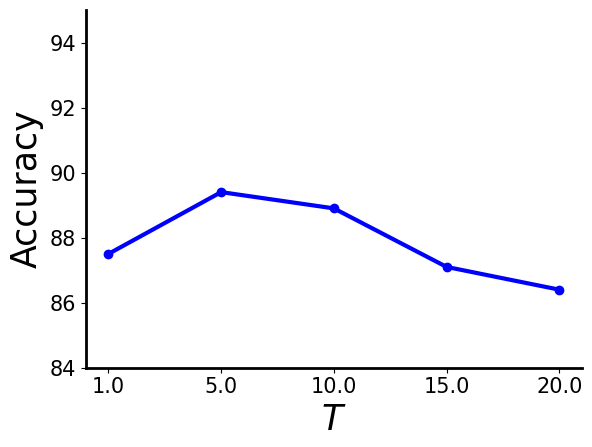}

\caption{$T$}
\end{subfigure}
\begin{subfigure}{0.23\textwidth}
\centering
\includegraphics[width = \textwidth, height=1.2in]{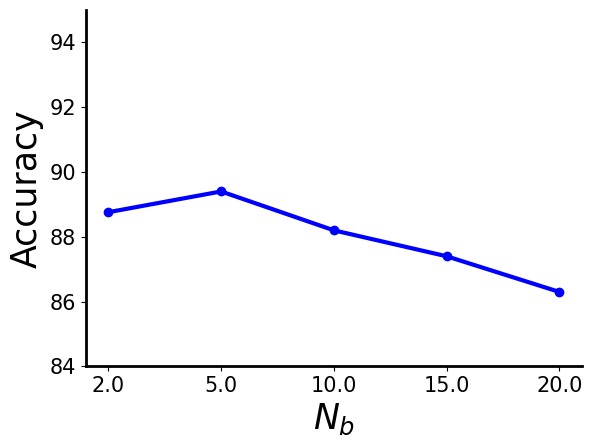}

\caption{$N_b$}
\end{subfigure}
	\caption{Sensitivity analysis for four hyper-parameters  $\lambda_{\text{adv}}$, $\lambda_{\text{mix}}$, $T$ and $N_b$ on USPS dataset.}

\label{fig:hyper_sen}
\end{figure}

\section{Hyper-parameter Sensitivity Analysis}
\label{section:hyper-params}
We conducted a sensitivity analysis for the proposed MPBM method across four hyperparameters: $\lambda_{\text{adv}}$—the trade-off parameter for adversarial training loss, $\lambda_{\text{mix}}$—the trade-off parameter for the mixup data classification loss term, $T$—the number of iterations for Stochastic Gradient Langevin Dynamics, and $N_b$—the number of samples used for mixup. We performed experiments on the digits training set and reported the accuracy on the USPS dataset, testing a range of different values for each of the four hyperparameters independently.

As shown in Figure \ref{fig:hyper_sen}, as $\lambda_{\text{adv}}$ increases from 0.2 to 0.5, the accuracy improves. Beyond this point, further increases in $\lambda_{\text{adv}}$ result in a slight decline in accuracy, indicating that the optimal value is around 0.5.
Similarly, the mixup loss coefficient, $\lambda_{\text{mix}}$, shows a trend where the accuracy rises with increasing $\lambda_{\text{mix}}$, reaching its highest point at around 0.5. Beyond this value, accuracy decreases, suggesting that overemphasis on the mixup loss may adversely affect performance.
For the number of Stochastic Gradient Langevin Dynamics steps, $T$, starting from $T = 1$, accuracy improves as $T$ increases, peaking at $T = 5$. However, further increases in $T$ lead to a decline in accuracy, indicating that excessive Stochastic Gradient Langevin Dynamics steps may result in diminishing returns or even negative effects.
The number of instances in the mixup, $N_b$, has a noticeable influence on model accuracy as well. Our observations show that performance peaks at $N_b = 5$ and then progressively decreases with higher values of $N_b$. This pattern suggests that while incorporating a moderate number of mixup instances can significantly enhance performance by promoting model generalization and robustness to overfitting, excessively high numbers of instances may counterproductively introduce noise into the learning process. This excess noise can overwhelm the model, leading to a degradation in the quality of the learned features and a subsequent drop in overall model accuracy. Therefore, careful calibration of $N_b$ is crucial to maintaining an optimal balance between diversity enhancement and noise control in model training.

\section{Conclusion}
In this paper, we introduced a novel method, Model-aware Parametric Batch-wise Mixup (MPBM), 
to address the challenge of single domain generalization. 
MPBM leverages a parametric mixup generator 
to generate synthetic instances 
by utilizing feature-wise attention mechanisms. 
Model-aware adversarial queries are generated with stochastic gradient Langevin dynamics
to guide the mixing up learning process. 
The generated mixup data is subsequently integrated with the training data in a progressive manner, 
diversifying the training set and improving the generalizability of the prediction model. 
Through extensive experiments on benchmark datasets for single domain generalization, including DomainNet, various digit datasets, and PACS benchmarks, we demonstrated that the proposed MPBM consistently outperforms the state-of-the-art techniques.

\bibliographystyle{ieeetr}
\bibliography{ref}

\end{document}